\documentclass[letterpaper,10pt,journal,twoside]{ieeetran}
\IEEEoverridecommandlockouts
\usepackage{algorithm}
\usepackage{algorithmic}
\usepackage{amsfonts}
\usepackage{amsmath}
\usepackage{amssymb}
\usepackage{array}
\usepackage{bm}
\usepackage{breakcites}
\usepackage{color}
\usepackage{comment}
\usepackage{float}
\usepackage{gensymb}
\usepackage{graphicx}
\usepackage{multirow}
\usepackage{pdfpages}
\usepackage{pgfplots}
\usepackage[caption=false,font=footnotesize,subrefformat=parens,labelformat=parens]{subfig}
\usepackage{textcomp}
\usepackage{tikz}
\usepackage{pgfplots}
\usepackage{wrapfig}
\usepackage{xcolor}
\usepackage{tabularx}
\usepackage{dblfloatfix}
\usepackage[normalem]{ulem}
\useunder{\uline}{\ul}{}

\PassOptionsToPackage{hyphens}{url}\usepackage[breaklinks=true,hidelinks=true]{hyperref}

\graphicspath{{images/}}

\title{CitDet: A Benchmark Dataset\\ for Citrus Fruit Detection}

\author{Jordan A. James$^{1}$, Heather K. Manching$^{2}$, Matthew R. Mattia$^{3}$, Kim D. Bowman$^{3}$,\\ 
Amanda M. Hulse-Kemp$^{4,2}$, and William J. Beksi$^{1}$%
\thanks{Manuscript received: April 10, 2024; Accepted: September 11, 2024.}
\thanks{This article was recommended for publication by Associate Editor P. V.
K. Borges and Editor C. Cadena Lerma upon evaluation of the reviewers'
comments. This work was supported by the United States Department of
Agriculture (USDA) under USDA-ARS CRIS \#6034-21000-018-000-D,
\#6066-21310-006-000-D, and USDA-ARS Non-Assistance Cooperative Agreement
\#6066-21310-005-061-S.
\textit{(Jordan A. James and Heather K. Manching contributed equally to this
work.)}
\textit{(Corresponding author: William J. Beksi}.)}
\thanks{$^{1}$The authors are with the Department of Computer Science and
Engineering, 
The University of Texas at Arlington, Arlington, TX, USA. 
Emails:
jaj9608@mavs.uta.edu,
william.beksi@uta.edu.
}
\thanks{$^{2}$The author is with the Department of Crop and Soil Sciences, 
North Carolina State University, Raleigh, NC, USA. 
Email:
hkmanchi@ncsu.edu.
}
\thanks{$^{3}$The authors are with the Subtropical Insects and Horticulture
Research Unit, USDA Agricultural Research Service, Ft. Pierce, FL, USA.
Emails:
matthew.mattia@usda.gov, 
kim.bowman@usda.gov 
}
\thanks{$^{4}$The author is with the Genomics and Bioinformatics Research Unit,
USDA Agricultural Research Service, Raleigh, NC, USA.
Email:
amanda.hulse-kemp@usda.gov.
}
\thanks{Digital Object Identifier (DOI): 10.1109/LRA.2024.3474473}
}

\begin{document}
\maketitle

\begin{abstract} 
In this letter, we present a new dataset to advance the state of the art in
detecting citrus fruit and accurately estimate yield on trees affected by the
Huanglongbing (HLB) disease in orchard environments via imaging. Despite the
fact that significant progress has been made in solving the fruit detection
problem, the lack of publicly available datasets has complicated direct
comparison of results. For instance, citrus detection has long been of interest
to the agricultural research community, yet there is an absence of work,
particularly involving public datasets of citrus affected by HLB. To address
this issue, we enhance state-of-the-art object detection methods for use in
typical orchard settings. Concretely, we provide high-resolution images of
citrus trees located in an area known to be highly affected by HLB, along with
high-quality bounding box annotations of citrus fruit. Fruit on both the trees
and the ground are labeled to allow for identification of fruit location, which
contributes to advancements in yield estimation and potential measure of HLB
impact via fruit drop. The dataset consists of over 32,000 bounding box
annotations for fruit instances contained in 579 high-resolution images. In
summary, our contributions are the following: (i) we introduce a novel dataset
along with baseline performance benchmarks on multiple contemporary object
detection algorithms, (ii) we show the ability to accurately capture fruit
location on tree or on ground, and finally (ii) we present a correlation of our
results with yield estimations. 
\end{abstract}


\section*{Multimedia Material}
The demonstration, dataset, and source code for citrus fruit detection and
counting can be found at
\href{https://robotic-vision-lab.github.io/citdet}{https://robotic-vision-lab.github.io/citdet}.

\section{Introduction}
\label{sec:introduction}
\begin{figure}[t]
\vspace{3mm}
\centering
\includegraphics[scale=0.1225]{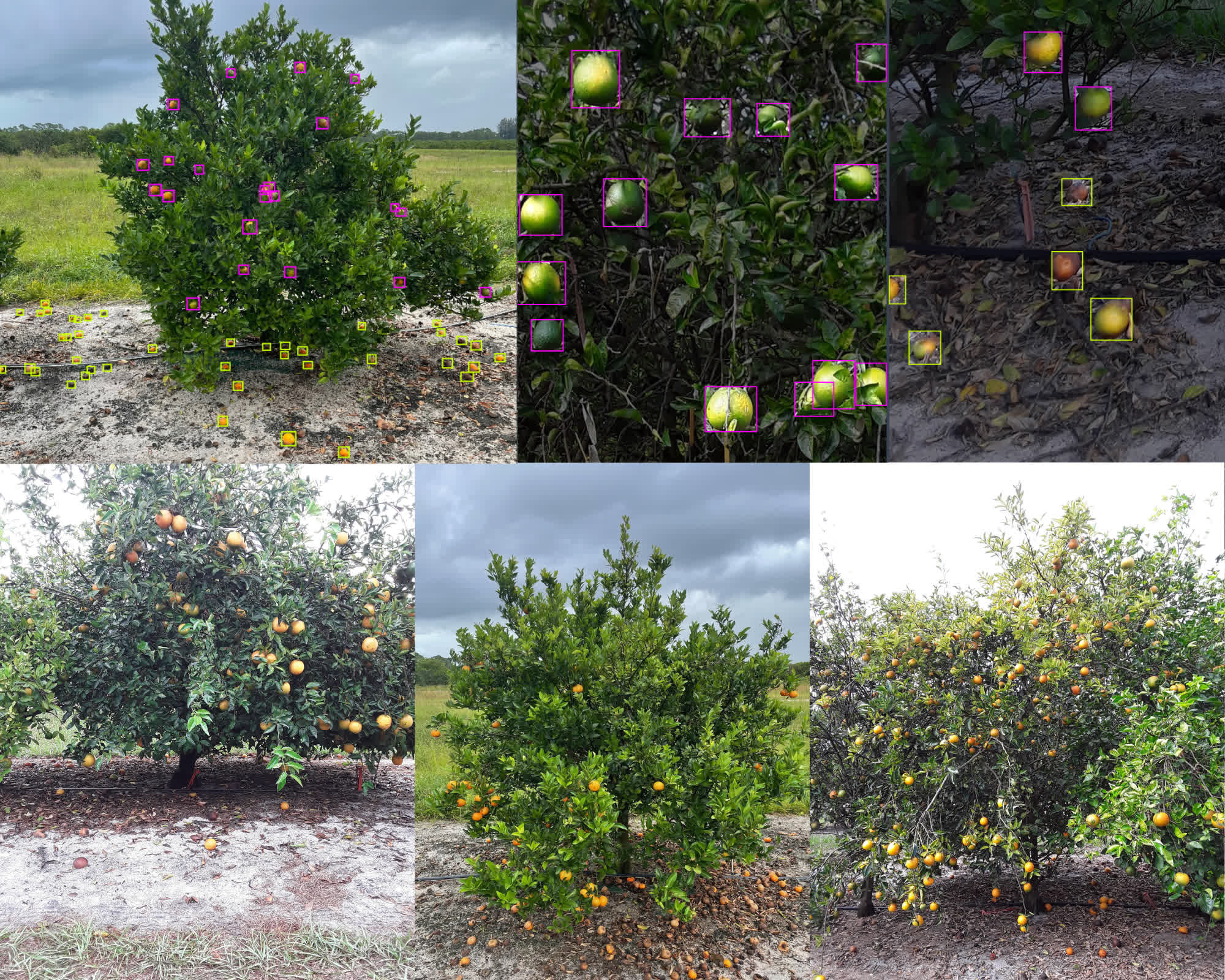}
\caption{The \textbf{CitDet} dataset contains precise bounding box object
annotations for fruit on tree (purple) and fruit on ground (yellow) (top row).
It also has images from multiple different tree rows including a large variety
of citrus (bottom row).}
\label{fig:1}
\end{figure} 

\IEEEPARstart{F}{ruit} detection and counting in orchards are crucial tasks for
agricultural automation. They can be used to reduce routine farming and
breeding activities as well as provide insightful estimates for harvest and
forthcoming growing seasons. Moreover, accurate fruit detection enables the
possibility of robotic harvesting, which has the potential to eliminate one of
the most labor-intensive processes for growers. Many imaging and sensing
technologies have been used for detecting fruit such as hyperspectral
\cite{ding2018feature}, laser scanning \cite{gene2019fruit}, thermal
\cite{gan2020active}, and RGBD sensors \cite{wang2017tree,lin2019field}, yet
the most common technology is the standard RGB camera. 

Although conventional RGB cameras are widely accessible, they present several
challenges for in-orchard fruit detection such as variation in appearance,
irregular lighting, and severe occlusion. Recent works have used deep learning
to overcome these difficulties \cite{koirala2019deep}. However, due to the lack
of standardized benchmark datasets for agricultural automation, it is difficult
to compare these methods with each other. Citrus datasets (e.g., CitrusFarm
\cite{teng2023multimodal}) have made large amounts of in-orchard citrus images
available, yet there still remains a lack of annotated citrus for detection. To
tackle this problem we establish a new benchmark dataset, \textbf{CitDet}
\cite{mavmatrix/dataset.2024.05.005,USDA.ADC/1529611}, for citrus fruit
detection and counting in orchard settings together with a comprehensive
analysis of state-of-the-art object detection algorithms, Fig.~\ref{fig:1}.

Computer vision techniques combined with deep learning are appealing in
agricultural automation due to their powerful prediction capabilities and
non-invasive nature. Nonetheless, such methods require huge amounts of data to
perform with high accuracy. While there exists large datasets (e.g., COCO
\cite{lin2014microsoft}) that have allowed for the development of new
algorithms, many automation tasks require custom datasets to achieve meaningful
results. For example, even though COCO contains a class for oranges, an orange
detector trained solely from the dataset instances will perform poorly in an
orchard setting. This is due to the complex background scenes in COCO, which
requires the detector to have more background classification machinery
resulting in less orange classification mechanisms \cite{miller2022object}.
Another reason for the weak performance stems from many instances of oranges
with very little or no occlusions in COCO. In orchard fields, oranges can be
occluded by other instances of oranges as well as the surrounding foliage.
Furthermore, orange instances will not be a large percentage of the image when
an entire tree is imaged. These pitfalls highlight the need for unique datasets
to carry out fruit detection, yield mapping, and much more. 

\subsection{Citrus Greening}
Citrus greening, also known as Huanglongbing (HLB), is a serious bacterial
disease that affects citrus trees and is threatening the existence of the
citrus industry, particularly in the southeastern United States
\cite{alvarez2016citrus}. HLB is transmitted by the Asian citrus psyllid, a
small insect that feeds on the leaves and shoots of citrus trees. Symptoms of
citrus greening include yellowing of the leaves, uneven ripening of the fruit,
and reduced fruit size and yield, both in the production of fruit as well as
premature fruit loss with fruits dropping to the ground. If left untreated, the
disease will ultimately lead to the death of the tree. Currently, there are no
effective treatments to cure trees affected by citrus greening. The best way to
control the spread of the disease is through a combination of methods such as
cultural, biological, chemical, and genetic control along with nutritional
management. Nevertheless, none of these countermeasures are foolproof. HLB
continues to spread and cause major economic losses for citrus growers.

\subsection{Challenges in Detecting Citrus Fruits in Native Orchard Environments}
While public datasets for citrus detection exist (e.g., OrangeSORT
\cite{zhang2022deep}), to the best of our knowledge \textbf{CitDet} is the
first dataset to include images of citrus trees affected by HLB, accompanied by
paired ground-truth metadata for fruit count. As fruit from an infected tree
are likely to be discolored and deformed, labeled examples of affected fruits
enhance the robustness of citrus detectors in orchard environments plagued by
the disease. Images in contemporary datasets are also low resolution, thus
limiting the amount of information to learn and predict from. Finally, the
aforementioned datasets only annotate fruit on trees. This can be an issue when
training a detector for yield estimation since the model will not learn the
difference between dropped fruit and fruit still on the tree, which can be
helpful in identifying symptoms of HLB or secondary fungus associated with post
bloom fruit drop. In contrast, \textbf{CitDet} aims to boost automation for
citrus orchards through high-resolution images of HLB infected citrus trees
with accurate annotations for both fruit on tree and fruit on ground. We
established an interdisciplinary team to address these challenges, including
subject matter experts in the biological system. Our contributions are as
follows.
\begin{itemize}
  \item We captured 579 high-resolution images of citrus trees affected by HLB.
  \item We provide over 32,000 high-quality annotations of in-orchard citrus,
  with separate classes of fruit, both on the ground and on the trees.
  \item We make these images, annotations, and associated fruit count metadata
  available in multiple public spaces to best serve the broader research
  communities. The \textbf{CitDet} dataset is available through MavMatrix
  \cite{mavmatrix/dataset.2024.05.005} and the USDA Ag Data Commons
  \cite{USDA.ADC/1529611}. 
  \item We analyze the baseline performance of detecting citrus fruits using
  multiple state-of-the-art object detection algorithms.
\end{itemize}

\section{Related Work}
\label{sec:related_work}
Benchmark datasets provide the means to train and evaluate new algorithms, they
are essential to the advancement of deep learning methods. Datasets such as
ImageNet \cite{deng2009imagenet}, Pascal VOC \cite{everingham2010pascal}, and
COCO \cite{lin2014microsoft}, have publicly released millions of labeled images
that contain a variety of classes and objects. Even though these datasets have
enabled breakthroughs in image classification, object detection, and object
segmentation, there is still a high demand for specialized datasets in
agricultural automation. Despite the fact that the number of agricultural
datasets continues to grow, there remains a lack of data on fruit affected by
common diseases.

\begin{table*}[b]
\centering
\caption{Comparison of fruit detection datasets}
\begin{tabular}{l|cccccc}
 Fruit Detection       & Fruit  & \# of Images & \# of Annotations & \# of Classes & Resolution  & Ground Truth \\ 
\hline
 Bargoti et al. \cite{bargoti2017deep}     & Apple  & 841   & 5,765  & 1 & 308 $\times$ 202    & Circles \\
 Stein et al. \cite{stein2016image}        & Mango  & 1,404 & 7,065  & 1 & 500 $\times$ 500    & Circles \\
 H{\"a}ni et al. \cite{hani2020minneapple} & Apple  & 1,001 & 41,325 & 1 & 720 $\times$ 1,280  & Polygons \\
 Hou et al. \cite{hou2022detection}        & Citrus & 4,855 & 17,567 & 1 & 1,280 $\times$ 720  & Boxes \\
 \textbf{CitDet}                           & Citrus & 579   & 44,233 & 2 & 2,448 $\times$ 3264 & Boxes \\
 \textbf{CitDet (Tiled)}                   & Citrus & 5,211 & 46,630 & 2 & 816 $\times$ 1,088  & Boxes    
\end{tabular}
\label{tab:1}
\end{table*}

\subsection{Fruit Detection}
Classical methods for fruit detection involve the use of static color
thresholds. These methods were enhanced by adding additional sensors such as
thermal and near-infrared cameras \cite{gongal2015sensors}. In recent years,
deep learning-based approaches have shown great promise in the domain of fruit
detection. For example, Bargoti and Underwood \cite{bargoti2017deep} and
H{\"a}ni et al. \cite{hani2020minneapple} used region-based convolutional
neural networks to detect fruits in images. Liu et al. \cite{liu2018robust}
utilized a fully-convolutional neural network to segment and detect fruits in a
tracking pipeline. DaSNet \cite{kang2019fruit} used atrous spatial pyramid
pooling and a gate feature pyramid network to detect and segment in-orchard
apples. Fast-FDM \cite{jia2022fast} leveraged FoveaBox \cite{kong2020foveabox}
with an EfficientNetV2 \cite{tan2021efficientnetv2} backbone to find green
apples in the field.

Many works use the YOLO \cite{redmon2016you} family of object detectors for
fruit tracking and yield estimation (e.g.,
\cite{mirhaji2021fruit,hou2022detection,zhang2022deep}). For instance, Wu et
al. \cite{wu2023detection} modified the YOLOv5 \cite{jocher2020yolov5} loss
function to detect banana bunches at harvest time via extended intersection
over union (IoU) \cite{peng2021systematic}. In addition, DeepLabV3+
\cite{chen2018encoder} was utilized to segment banana bunches for counting
during sterile bud removal time. Faster R-CNN \cite{ren2015faster}, and the
instance segmentation variant Mask R-CNN \cite{he2017mask}, have also been
successfully employed. Chu et al. \cite{chu2021deep} added a feature
suppression network to the end of Mask R-CNN to filter unwanted foliage from
the masks. Even though deep learning is advantageous for fruit detection, the
task remains challenging due to the unstructured background prevalent in
orchards. Tang et al. \cite{tang2023optimization} provided an extensive review
on strategies involving sensor, image processing, and algorithm optimization
for orchard environments.

In-orchard detection of fruits is vital for agricultural tasks including fruit
yield estimation and automated picking. For example, Dorj et al.
\cite{dorj2017yield} leveraged image processing, and the LAB color space, to
detect oranges and estimate yield when compared against human observation.
Faster R-CNN was used to detect citrus for automated size estimation using
drones \cite{apolo2020deep}. Behera et al. \cite{behera2021fruits} also
utilized Faster R-CNN for detection and yield estimation of various fruits.
YOLOv4 \cite{bochkovskiy2020yolov4} was deployed to detect apples and the
trunks of trees by Gao et al. \cite{gao2022novel}. The detections are used
downstream for apple counting by tracking of the tree trunks in videos. The
estimation of tree-level yield by Vijayakumar et al. \cite{vijayakumar2023tree}
involved the integration of citrus fruit detections from YOLOv3, hyperspectral
data, and tree-level traits as inputs for classical machine learning
algorithms. Wang et al. \cite{wang2022study} used YOLOv5 to detect individual
lychee and Mask R-CNN to segment the fruit and determine point locations for
picking.

\subsection{Comparison of Datasets}
Table~\ref{tab:1} highlights the issues with current fruit detection datasets.
Due to the high cost and time involved in labeling data, researchers have opted
to use small datasets resulting in a lack of variety. Many works attempt to
remedy this by tiling images into smaller sections to artificially increase the
overall number of images \cite{bargoti2017deep}. Although this technique
expands the dataset size, it only shows a small part of the original image and
it does not change the total number of object instances. Furthermore, the
diversity of the dataset does not increase, which can lead to overfitting when
developing machine learning models. While recent research has made available
many labeled citrus images \cite{hou2022detection}, the images fail to capture
the entire tree and they can suffer from similar overfitting problems. 

Another issue is that many of the datasets (e.g., MinneApple
\cite{hani2020minneapple}) contain fruit on the ground without any labels. This
can lead to ambiguity when training a model as it must learn not to detect the
fruit on the ground, which can have similar features to that of the fruit on
the tree. Other works attempt to solve this issue by treating the unlabeled
fruit as a ``no object" class and through the introduction of new loss terms to
penalize false positive predictions \cite{hou2022detection}. Our dataset aims
to fix these issues and provide researchers with a means to evaluate and
compare their algorithms. To the best of our knowledge, \textbf{CitDet} is the
\textit{first} fruit detection dataset to provide full-resolution images with
annotations for both fruit on trees and on the ground. It is the \textit{only}
citrus dataset to contain images of entire citrus trees. In addition, we
include a variety of citrus tree species imaged at different stages of maturity
under varying illumination conditions to avoid overfitting. 

\section{Image Collection}
\label{sec:image_collection}
\textbf{CitDet} is composed of images captured at the USDA Agricultural
Research Service Subtropical Insects and Horticulture Research Unit in Fort
Pierce, Florida, between October 2021 and October 2022. The orchard contains a
large assortment of citrus tree species and is used for genomics and
phenotyping research involving citrus infected with HLB. We collected 579
images from different sections of the orchard using Field Book
\cite{rife2014field} on Android tablets. While imaging the plants, we faced the
camera in a portrait orientation directly centered on the tree of interest. All
images were taken at the edge of the soil in the tree row to simulate a
ground-based robot imaging the tree while moving between two rows of trees. 

Data was collected over the course of one year to allow the fruit to be imaged
at different stages of the ripening cycle during citrus production from October
to March. Due to the nature of HLB and the mixture of fruit maturity,
\textbf{CitDet} contains fruit of different colors and sizes. All trees were
imaged from both the sunny and shady sides of the tree row and imaging was done
over multiple days to account for variations in lighting and weather
conditions. The imaged trees consist of 60 different varieties of citrus,
representing 8 different species of citrus and other related genera. All images
include a timestamp, tree ID, and the side of the tree imaged in the file name.

The dataset was split into 80\% for the training set and 20\% for the test set.
Images included in the test set were chosen to cover a variety of citrus tree
species and cultivars with fruits of varied color, shape, and size.
Specifically, images in the test set contain unique views of fruit imaged from
both shady and sunny sides of the tree at varied times of the day.
Additionally, we collected observed yield estimates for 187 randomly selected
trees by counting the number of fruit per tree prior to imaging. Out of the
acquired images 130 were not included in the final dataset. This was done to
avoid blurry and background fruit, which can be ambiguous to annotate.
Succeeding work may consider coarse annotations for fruit in the background.
Since we do not include a benchmark or dataset for tracking, we use the
predicted detections from the model as yield estimation and validate against
the ground truth.

\section{Image Annotation}
\label{sec:image_annotation}
To construct \textbf{CitDet}, we followed standard annotation methods found in
other benchmark object detection datasets. The Roboflow
\cite{dwyer2022roboflow} annotation tool was used to label the image data with
bounding boxes. Specifically, we labeled all visible fruit in the foreground
regardless of occlusions, while fruit in the background was only labeled for
instances with no occlusions. We do not provide any annotations for fruits that
are fully occluded and thus not visible. Each of the labeled objects in an
image are classified as either fruit on tree or fruit on ground. 

\textbf{CitDet} is unique for fruit detection in that we provide labels for
fruit that has already fallen. This allows detection methods to accurately
identify fruit that contributes to the overall fruit yield. Since a single
image can contain a large number of object instances and a heavy amount of
occlusions, labeling images is a time-consuming task. For instance, annotating
a single image can take up to half an hour. Therefore, the task of instance
labeling was split among several internal volunteers and each image was
assigned to only one volunteer to annotate. After labeling, each image was then
reviewed by a single expert to validate the quality of the annotations.

\section{Dataset Statistics}
\label{sec:dataset_statistic}
Popular image datasets (e.g., ImageNet, Pascal VOC, and COCO) focus on
detecting many categories of objects in a multitude of natural settings. In
each of these datasets, most object instances tend to occupy over 10\% of the
image area. As a result, the majority of images contain few object instances
per image. In comparison, \textbf{CitDet} provides a unique collection of
small, highly-cluttered objects under various levels of occlusion.

\begin{figure}
\includegraphics[width=\linewidth]{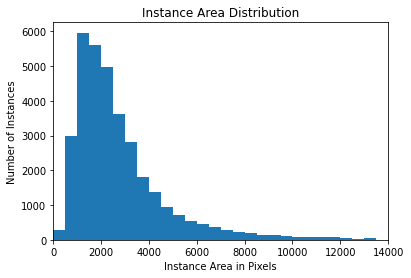}
\caption{The area distribution of the objects in \textbf{CitDet}. The dataset
is mainly comprised of small object instances with an area of less than 50$^2$
pixels.}
\label{fig:2}
\end{figure}

Since the \textbf{CitDet} dataset does not contain many different categories,
it is comparable to the Caltech Pedestrian Detection
\cite{dollar2009pedestrian}, KITTI \cite{geiger2013vision}, and MinneApple
datasets where each dataset focuses on providing many labeled object instances
per category. Specifically, our dataset is similar to MinneApple as both
provide a large number of annotations for fruit in an orchard setting. The main
difference is that our dataset provides annotations of fallen fruit as well as
fruit still on the tree. \textbf{CitDet} allows models to automatically
distinguish between dropped and non-dropped fruit, enabling accurate yield
estimation without any additional postprocessing. 

\begin{figure}
\includegraphics[width=\linewidth]{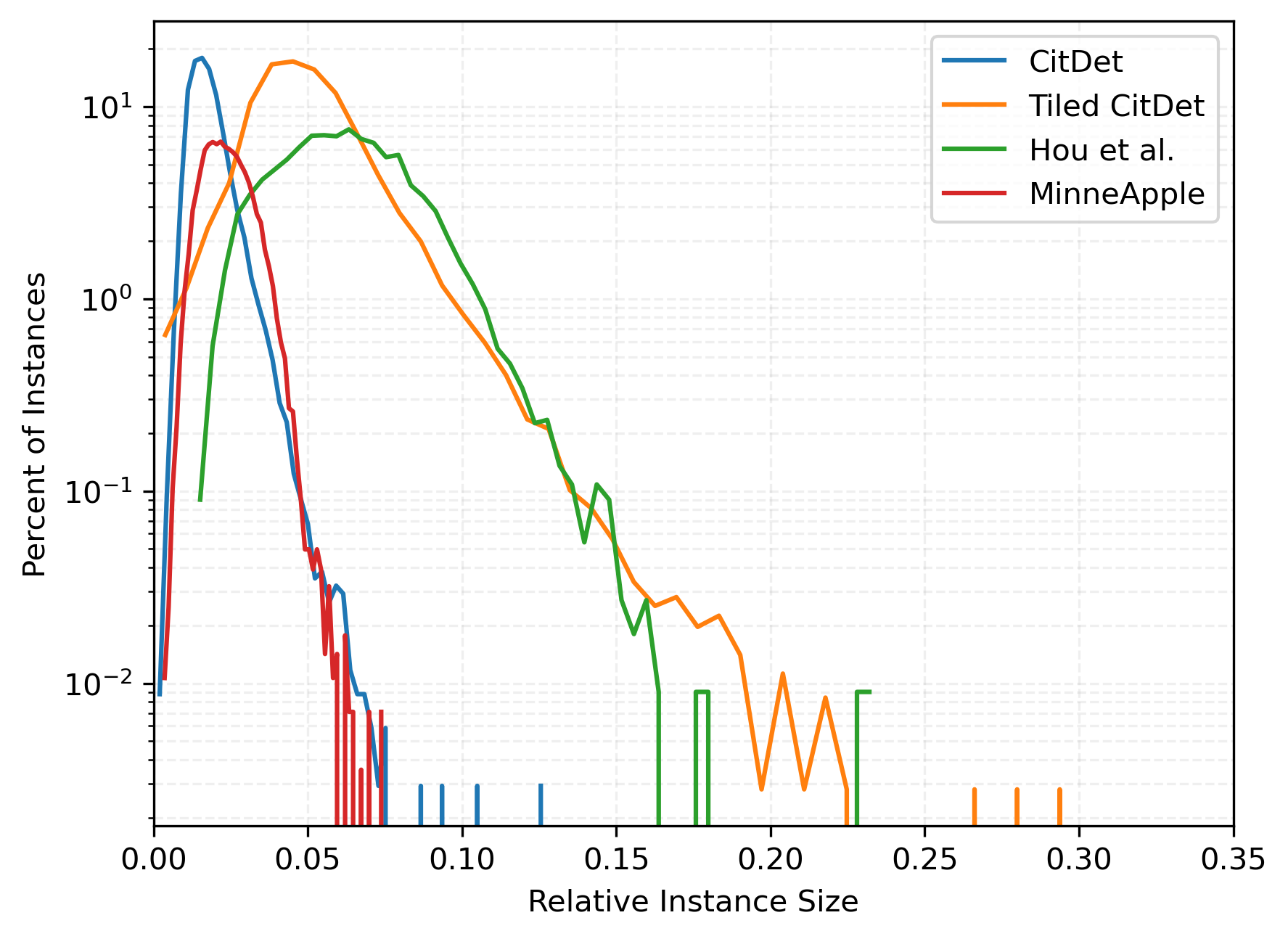}
\caption{A comparison of the relative instance size (square root of instance
area divided by total image area) between fruit detection datasets.}
\label{fig:22}
\end{figure}

\begin{figure}
\includegraphics[width=\linewidth]{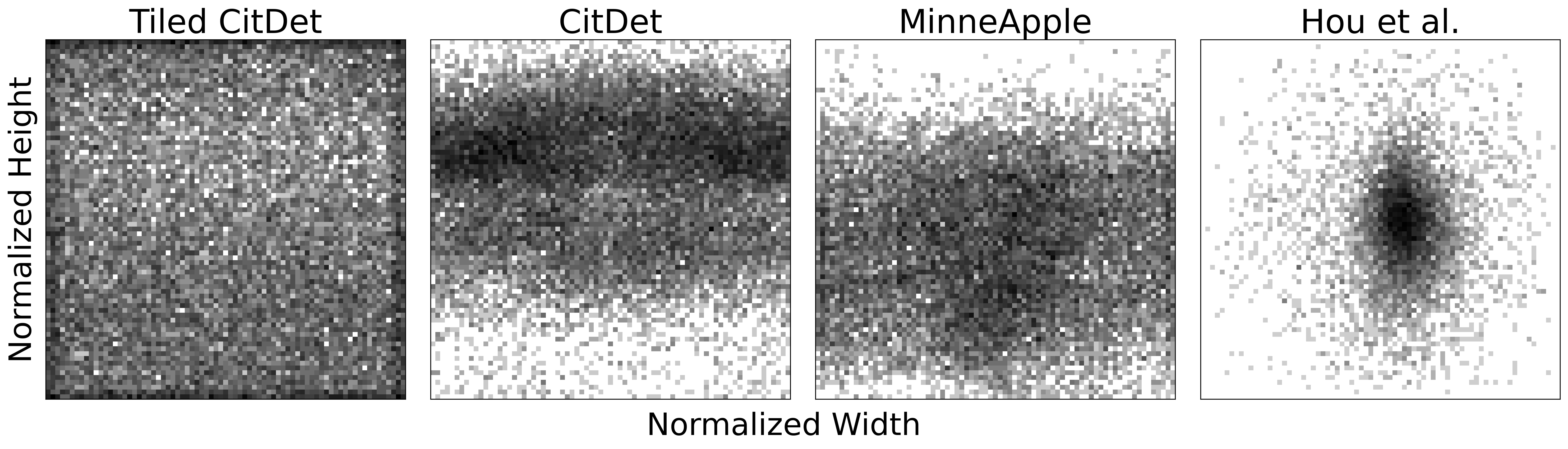}
\caption{The distribution of object centers in normalized image coordinates for
four datasets. \textbf{CitDet} exhibits the greatest spatial diversity among
the citrus datasets, with the tiled version exhibiting greater complexity than
all the other fruit datasets.}
\label{fig:21}
\end{figure}

\textbf{CitDet} also provides a challenge for state-of-the-art detection
algorithms due to the small average size of the object instances. Small objects
are generally harder to detect and require specialized network structures. In
ImageNet, Pascal VOC, and COCO, only 50\% of the objects occupy less than 10\%
of an image while the remaining 50\% of the objects occupy between 10\% and
100\% of an image. Conversely, over 90\% of the objects in \textbf{CitDet} take
up less than 10\% of an image. As displayed in Fig.~\ref{fig:2}, the average
object in our dataset is only 50 $\times$ 50 pixels, which corresponds to less
than 0.1\% of an image. In contrast, objects in MinneApple occupy around 0.17\%
of an image. Although the average object in \textbf{CitDet} is smaller, the
distribution of relative instance size is similar to MinneApple as shown in
Fig.~\ref{fig:22}. The high resolution of \textbf{CitDet} allows for
preprocessing techniques (e.g., tiling) to shift the relative distribution and
increase spatial complexity. This increases the versatility of the dataset as
highlighted in Fig.~\ref{fig:21}. Another key difference between
\textbf{CitDet} and MinneApple is the format of the object annotations,
bounding boxes versus polygons, respectively. 

\section{Algorithmic Analysis}
\label{sec:algorithmic_analysis}
We applied multiple state-of-the-art object detection algorithms on
\textbf{CitDet} to establish a baseline for future work. For all the
experiments, we used the same 460 images from our dataset for training while
the remaining 119 images were utilized as the test set. We divided the
experiments into two tasks: (i) detection on the whole (full-resolution) image
dataset, and (ii) detection on the tiled image dataset. The tiled image dataset
was created by splitting every image into 9 separate images using a 3 $\times$
3 cropping method. For any bounding box split by the cropping, a bounding box
for each portion of the object was created in each of the new images resulting
in slightly more object instances. The resulting dataset contains 5,211 images
at a resolution of 816 $\times$ 1088.

\begin{figure*}
\centering
\subfloat[Faster R-CNN \label{frcnn_example}]{\includegraphics[scale=0.0675]{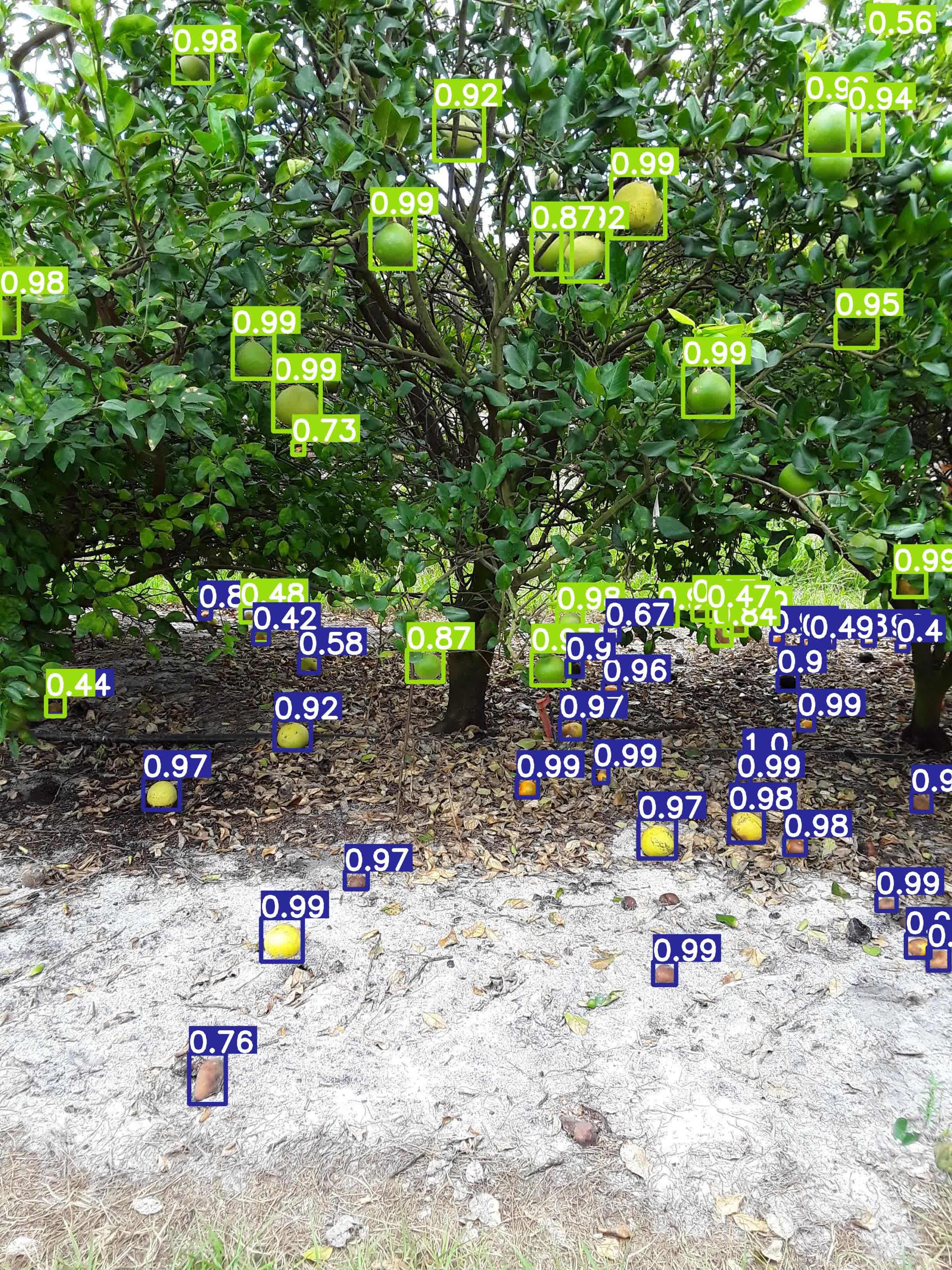}}
\hspace{0.1em}
\subfloat[YOLOv5 \label{yolov5_example}]{\includegraphics[scale=0.0675]{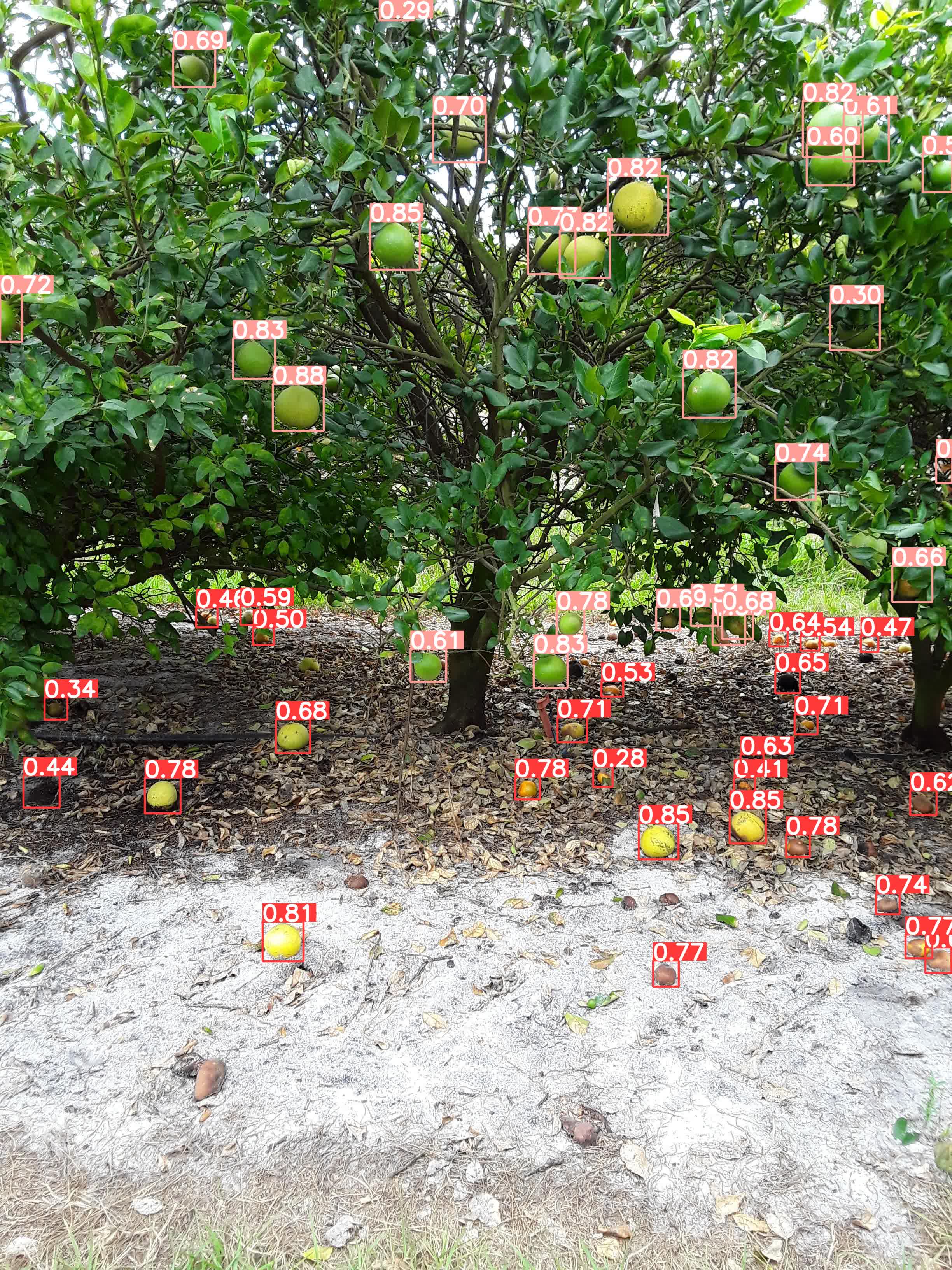}}
\hspace{0.1em}
\subfloat[YOLOv7 \label{yolov7_example}]{\includegraphics[scale=0.0675]{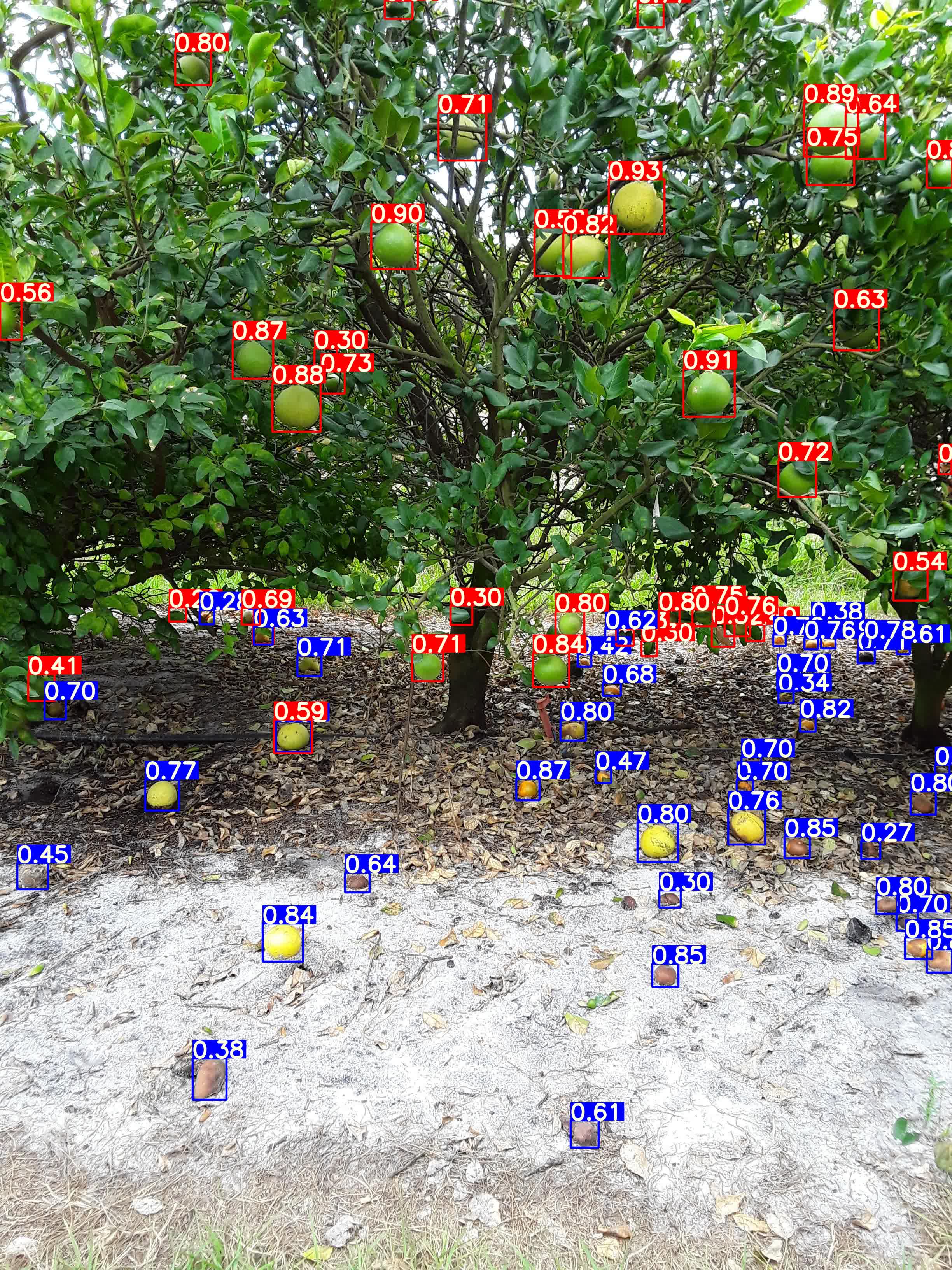}}
\caption{Visualizations of the qualitative results for each network on the same
full validation image. The bounding box annotations are colored based on the
predicted class label with differing color schemes for each network.}
\label{fig:3}
\end{figure*}

\begin{figure*}
\centering
\subfloat[Faster R-CNN \label{frcnnt_example}]{\includegraphics[scale=0.185]{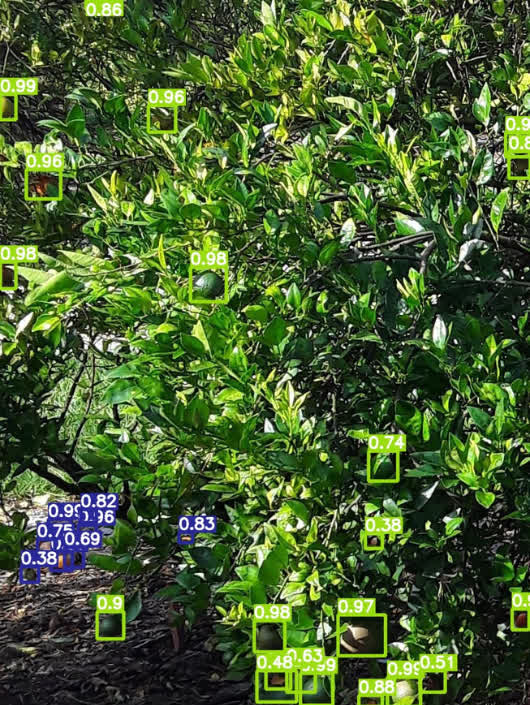}}
\hspace{0.025em}
\subfloat[YOLOv5 \label{yolov5t_example}]{\includegraphics[scale=0.185]{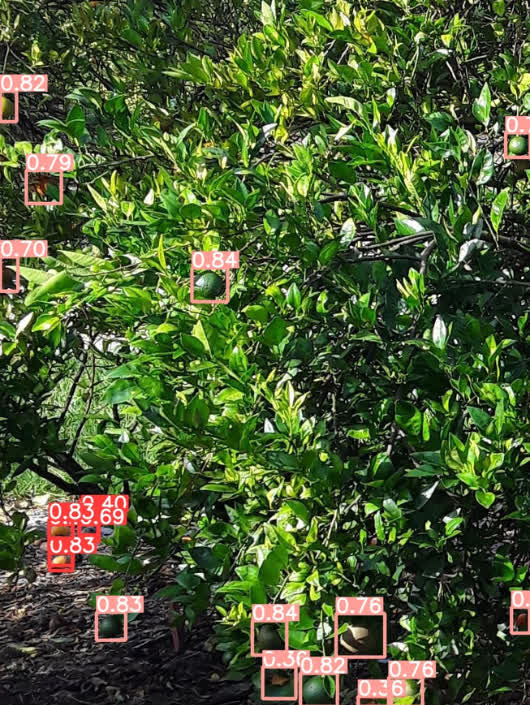}}
\hspace{0.025em}
\subfloat[YOLOv7 \label{yolov7t_example}]{\includegraphics[scale=0.185]{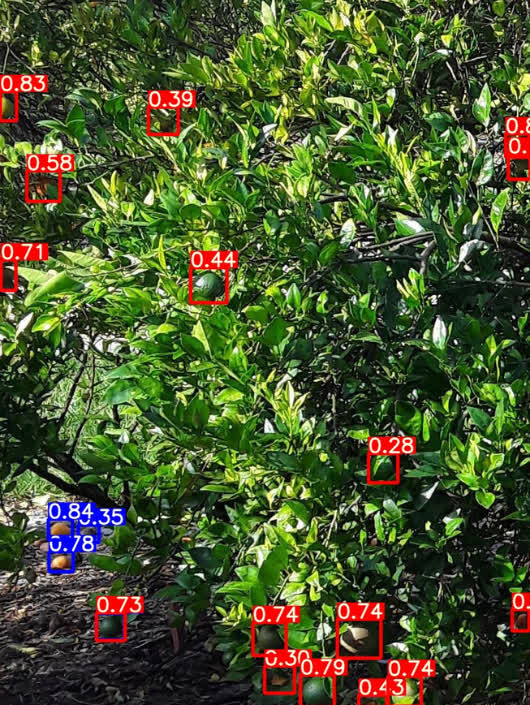}}
\hspace{0.025em}
\subfloat[DETR \label{detr_example}]{\includegraphics[scale=0.185]{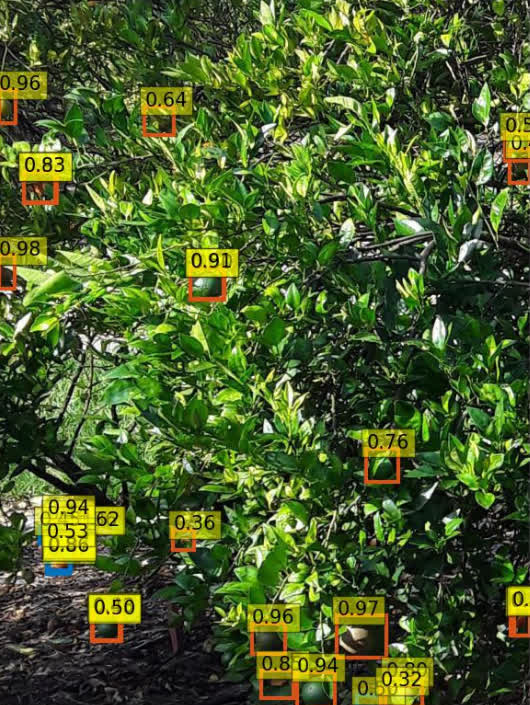}}
\hspace{0.025em}
\subfloat[YOLOS \label{yolos_example}]{\includegraphics[scale=0.185]{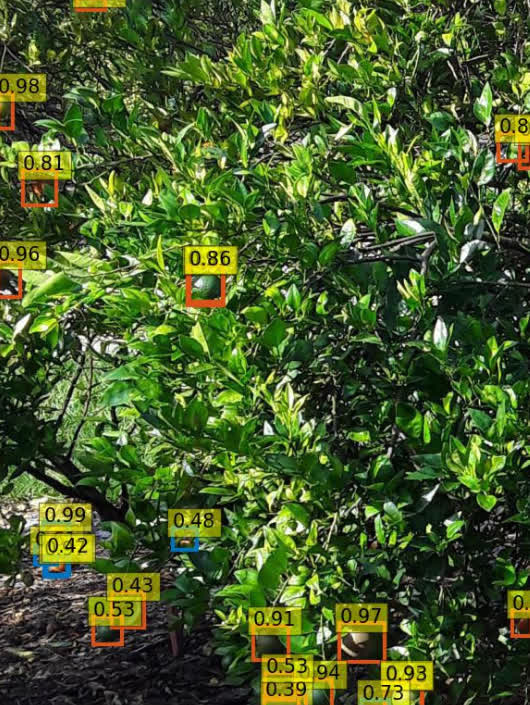}}
\caption{Visualizations of the qualitative results for each network on the same
tiled validation image. The bounding box annotations are colored based on the
predicted class label with differing color schemes for each network.}
\label{fig:4}
\end{figure*}

\subsection{Fruit Detection}
In all the experiments, the models were trained using an NVIDIA Tesla T4 GPU
and restricted to similar levels of compute power. Every algorithm tested used
pretrained COCO weights for initialization and they were fine-tuned for 50
epochs on the tiled dataset. The models tested on the whole image dataset used
pretrained weights from the tiled dataset and they were fine-tuned for 50
epochs on resized images of 704 $\times$ 704 resolution. We evaluated each of
the algorithms on the test set and reported the average performance for each
class as well as the mean average performance of all classes.

\textbf{Detection evaluation metrics:} We followed the established evaluation
protocols used by other benchmark object detection datasets (e.g.,
\cite{everingham2010pascal,lin2014microsoft,hani2020minneapple}). For a single
class, we reported the average precision (AP) as the main evaluation metric. We
also recorded the mean AP (mAP) of all classes. For the AP of a class we used
an IoU threshold of 0.5, increased it in intervals of 0.05 up to 0.95, and
reported the mean of the AP at each threshold. The AP at thresholds of 0.5 and
0.75, as well as the AP for small (AP$_S$), medium (AP$_M$), and large (AP$_L$)
objects, are also provided.

\textbf{Faster R-CNN:} Faster R-CNN utilizes a ResNet-50 \cite{he2016deep}
backbone along with a feature pyramid network. The implementation makes use of
region proposal networks as the head of the detection network and two branches
for predicting the class and bounding boxes. We used a k-means clustering
algorithm with IoU and ground-truth labels in the training dataset to determine
the anchor box sizes. All the other optimization parameters were the same as
the original paper.

\textbf{YOLOv5:} The implementation of YOLOv5 is provided by Ultralytics.
YOLOv5 uses a different architecture compared to previous versions of YOLO. In
particular, it has a cross stage partial backbone architecture, which is a
combination of two types of convolutional layers: depth-wise convolution and
point-wise convolution. The architecture not only reduces the computational and
memory requirements of the model, but also improves the overall accuracy.

\begin{table}
\centering
\caption{Whole Image Dataset Results}
\begin{tabular}{cccccccc} 
\hline
Model        & \# of Params & AP            & AP$_{50}$     & AP$_{S}$      & AP$_{M}$      & AP$_{L}$ \\ \hline
Faster R-CNN & 41.3 M       & 22.0          & 51.5          & \textbf{21.8} & \textbf{44.2} &  - \\
YOLOv5       & 21.1 M       & 34.8          & 70.0          & 9.0           & 36.9          & 46.3 \\
YOLOv7       & 36.5 M       & \textbf{40.6} & \textbf{77.9} & 12.3          & 42.6          & \textbf{49.7} \\ 
\hline
\end{tabular}
\label{tab:2}
\end{table}

\begin{table}
\centering
\caption{Tiled Image Dataset Results}
\begin{tabular}{cccccccc} 
\hline
Model        & \# of Params & AP            & AP$_{50}$     & AP$_{S}$      & AP$_{M}$      & AP$_{L}$ \\ 
\hline
YOLOS        & 30.9 M       & 32.4          & 70.7          & 10.3          & 35.8          & 47.9 \\
DETR         & 41.3 M       & 35.0          & 72.8          & 9.8           & 38.7          & 49.8 \\
\hline
Faster R-CNN & 41.3 M       & 37.2          & 76.0          & \textbf{24.1} & 44.4          & 45.9 \\
YOLOv5       & 21.1 M       & 44.9          & 81.9          & 18.2          & 49.2          & 50.6 \\
YOLOv7       & 36.5 M       & \textbf{45.5} & \textbf{83.1} & 19.2          & \textbf{49.4} & \textbf{52.1} \\ 
\hline
\end{tabular}
\label{tab:3}
\end{table}

\textbf{YOLOv7:} YOLOv7 \cite{wang2023yolov7} improves upon real-time object
detection performance by designing a trainable ``bag-of-freebies" set of
methods. The detector addresses two new issues in object detection evolution by
proposing the ``extend" and ``compound scaling" techniques. The proposed
methods reduce the number of parameters and the amount of computation by
approximately 40\% and 50\%, respectively. They also result in faster inference
speed and higher detection accuracy.

\textbf{DETR:} The DETR \cite{carion2020end} model for object detection was
introduced by Facebook Artificial Intelligence Research. It uses a ResNet-50
backbone to embed image features as input to a transformer encoder-decoder. The
architecture utilizes a standard transformer encoder-decoder with the addition
of object query tokens passed to the decoder. The network is trained using a
bipartite matching loss, thus removing the need for non-maximum suppression. We
used sine positional embeddings for both the encoder and decoder.

\textbf{YOLOS:} You Only Look at One Sequence (YOLOS) \cite{fang2021you}
leverages a vanilla vision transformer (ViT) \cite{dosovitskiy2021image}
encoder with minimal multilayer perceptron heads for object detection. The
YOLOS ViT architecture is the same as DeiT \cite{touvron2021training} and
includes a distillation token for training. Images are represented by a
sequence of 16 $\times$ 16 patches. The patches are projected and inputted as
tokens to the transformer encoder along with positional embedding tokens. YOLOS
feeds detection token inputs to an encoder, which are then decoded by the head
to predict the bounding box and class. Similar to DETR, the network is trained
using a bipartite matching loss.  We utilized sine positional embeddings for
the implementation tested.

\subsection{Discussion}
Contrary to other fruit detection works, we found that processing the image in
a tiled manner performs better than detectors that operate on the full image.
We hypothesize that there are two primary factors for this. The first factor is
the difference between the tiling methods used. For example, in the Bargoti and
Underwood \cite{bargoti2017deep} model, training images are split into
overlapping chunks, which requires the filtering of overlapping bounding boxes.
Our tiling method does not have any overlap. Instead, we process the bounding
boxes on the edges of the fruits before training and therefore do not require
the filtering step at the end. The second factor contributing to the improved
performance of the tiled versions is the ability to preserve information when
resizing the image for the model. 

In comparing the state-of-the-art object detection algorithms, we found that
YOLOv7 outperformed all the other algorithms on both the whole and tiled
images. However, Faster R-CNN performed the best for small objects on both
image types. The results in Tables~\ref{tab:2} and \ref{tab:3} suggest that the
YOLO architectures benefit more from transfer learning on high-resolution
images when compared to Faster R-CNN. All the networks demonstrate the
capability to distinguish between fruit yield and drop even in ambiguous cases
such as in Fig.~\ref{fig:3} and Fig.~\ref{fig:4}. We observed that all the
methods struggled to accurately detect small object instances. We note that
standard convolutional approaches performed better than transformers for small
object detection. Nonetheless, future research is needed to accurately detect
small and medium sized objects for all methods. Our findings confirm previous
works (e.g., \cite{hoiem2012diagnosing,hani2020minneapple}), which found that
object size is one of the main determinants of error in object detection. 

\subsection{Yield Estimation}
Yield estimation is performed at the tree level to provide higher-resolution
information for downstream tasks such as phenomic selection for citrus
breeding. Additional analysis at the row level can be performed by aggregation
of the tree-level estimates. To further validate the detection results, we used
a detection only based yield estimation method for each tree and compared it
with the ground-truth estimates as currently used in breeding programs for
observed fruit yield values. In this experiment, we acquired front and back
images of 187 trees before harvest. A YOLOv7 model was trained on a subset of
\textbf{CitDet}, which excluded images of trees in the yield estimation test
set. Breeding program measured fruit yield estimates for each tree were
recorded and used as the ground-truth yield values. 

For all the experiments we consider only the fruit on the trees for prediction
of the fruit yield. We evaluated the per-tree yield prediction correlation with
the ground-truth estimates using the square of the Pearson correlation
coefficient,
\begin{equation*} \tag{1} 
  R^2 =
  \bigg(\frac{ \sum_{i=1}^{n}(x_i-\bar{x})(y_i-\bar{y}) }{%
        \sqrt{\sum_{i=1}^{n}(x_i-\bar{x})^2}\sqrt{\sum_{i=1}^{n}(y_i-\bar{y})^2})}\bigg)^2,
  \label{eq:1}
\end{equation*}
where $n$ is the total number of trees in the experiment, $x$ and $y$ are the
predicted and ground-truth yield estimates, and $\bar{x}$ and $\bar{y}$ are the
mean of the predicted and ground-truth yield estimates. Two simple yield
estimation methods were compared: detect-count and filter-detect-count. The
R$^2$ values are reported in Table~\ref{tab:4}.

\begin{table}
\centering
\caption{Tree Yield Correlation}
\begin{tabular}{cccc} 
\hline
Method & Front R$^2$ & Back R$^2$ & Front+Back R$^2$ \\ 
\hline
detect-count           & 0.584         & 0.635       & 0.754 \\
filter-detect-count    & 0.603         & 0.677       & 0.793 \\
\hline
\end{tabular}
\label{tab:4}
\end{table}

The first method for yield estimation, detect-count, simply detects the fruit
in an image and estimates the counts of fruit labeled as fruit on the tree with
a confidence of at least 0.3. Detection is applied to both the front and back
images, and yield is estimated separately for each. We made the assumption that
due to the high amount of foliage on the trees, fruits that are visible in one
view will be occluded in the other view. As such, the sum of the predicted
counts for each view of the tree are used as the predicted fruit yield. 

\begin{figure}
\centering
\includegraphics[scale=0.27]{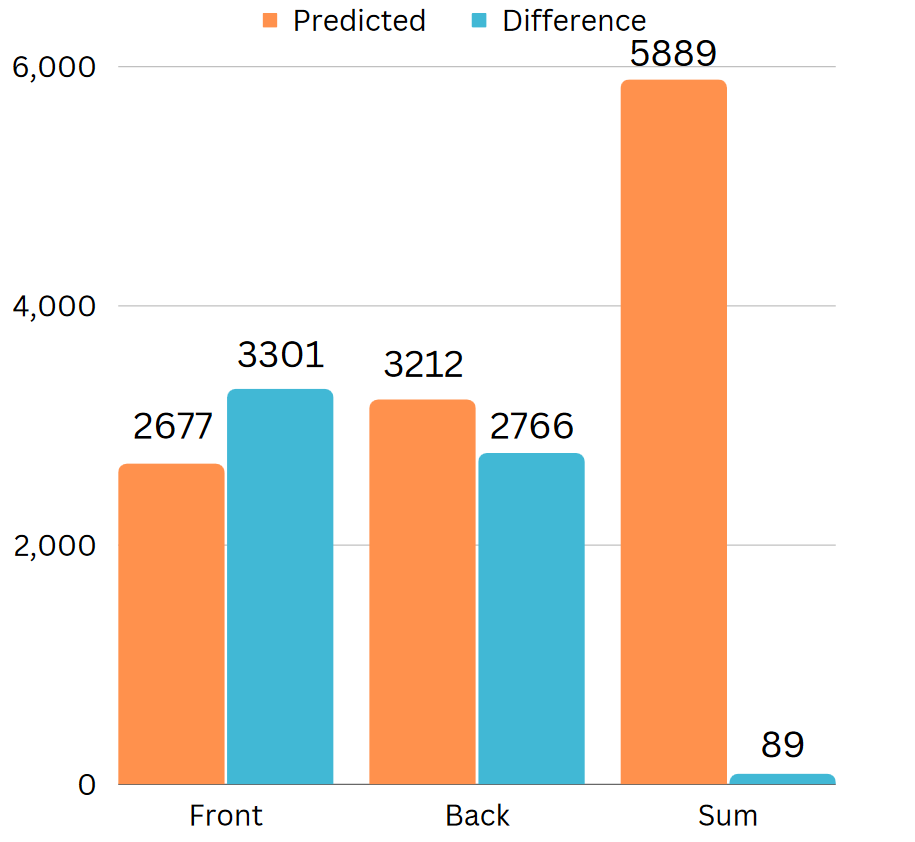}
\caption{The total predicted yield and the difference of actual and predicted
yield for 187 sample citrus trees of varying species and maturity obtained
using the filter-detect-count method. We used infield hand counts acquired by
technicians as the ground-truth yield.}
\label{fig:5}
\end{figure}

An analysis of large prediction errors showed that many images contained parts
of neighboring trees in the image, which attributed to higher fruit counts. The
second method, filter-detect-count, aims to address this issue by using a
two-stage detection approach to filter out neighboring trees. The first stage
uses a separate model to detect all trees and their canopies. Then, the central
most detection is taken as input to the second stage and passed to a YOLO model
for fruit detection. The improved method correctly predicted four out of the
five top-producing trees with an error of only -1.49\% on the total yield
estimation of the orchard as shown in Fig.~\ref{fig:5}.

\section{Conclusion and Future Work} 
\label{sec:conclusion_and_future_work}
In this work we introduced \textbf{CitDet}, a new dataset for citrus detection
in orchards prone to citrus greening. By distinguishing between fruit on the
ground and fruit on the trees, we showed that it is possible to identify fruit
that contributes to yield via a single model. With this annotated collection of
citrus, we aim to increase the comparability of fruit detection algorithms and
advance research on this topic. Importantly, the metadata and associated
tree-level yield estimates are included to serve as a ground-truth metric. 

Detectors trained on \textbf{CitDet} can be applied for citrus tracking and
yield estimation, size estimation, and automated fruit picking in follow-up
studies. The results of the object detection benchmarks indicate that
\textbf{CitDet} is challenging for modern detectors and there are many
opportunities for subsequent research and improvement. In particular, future
work is necessary to improve the accuracy of detectors on small objects in
cluttered environments and the inclusion of biological ground-truth data will
allow for further analysis.

\section*{Acknowledgments}
The authors thank members of the Hulse-Kemp Lab including Dr. Keo Corak, Dr.
Emily Delorean, Ashley Schoonmaker, Grant Billings, and Cassie Newman for
assistance in image annotation and USDA-ARS USHRL technicians Jefferson Shaw
and Darren Cole for assistance in image collection. 

\bibliographystyle{IEEEtran}
\bibliography{IEEEabrv,citdet_a_benchmark_dataset_for_citrus_fruit_detection}   
\end{document}